\documentclass[letterpaper, 10 pt, conference]{ieeeconf}  % Comment this line out if you need a4paper
\IEEEoverridecommandlockouts
\usepackage{cite}
\usepackage{amsmath,amssymb,amsfonts}
\usepackage{algorithmic}
\usepackage{graphicx}
\usepackage{textcomp}
\usepackage{xcolor}
\usepackage{threeparttable}
\DeclareMathOperator{\E}{\mathbb{E}}

\usepackage{amsmath}
\usepackage{amsfonts}
\usepackage{amssymb}
\usepackage{array}
\usepackage{tabu}
\usepackage{graphicx}
\usepackage{algorithm}
\usepackage{algorithmic}
\usepackage{wrapfig}
\usepackage{comment}

\def\BibTeX{{\rm B\kern-.05em{\sc i\kern-.025em b}\kern-.08em
    T\kern-.1667em\lower.7ex\hbox{E}\kern-.125emX}}
    
\title{\LARGE \bf SAFARI: Safe and Active Robot Imitation Learning with Imagination}
\author{Norman Di Palo and Edward Johns% <-this % stops a space

\thanks{The Robot Learning Lab at Imperial College London, London, UK. 
        \tt\small n.di-palo20@imperial.ac.uk}}%

\begin{document}

\maketitle

\begin{abstract}
     One of the main issues in Imitation Learning is the erroneous behavior of an agent when facing out-of-distribution situations, not covered by the set of demonstrations given by the expert. In this work, we tackle this problem by introducing a novel active learning and control algorithm, SAFARI. During training, it allows an agent to request further human demonstrations when these out-of-distribution situations are met. At deployment, it combines model-free acting using behavioural cloning with model-based planning to reduce state-distribution shift, using future state reconstruction as a test for state familiarity. We empirically demonstrate how this method increases the performance on a set of manipulation tasks with respect to passive Imitation Learning, by gathering more informative demonstrations and by minimizing state-distribution shift at test time. We also show how this method enables the agent to autonomously predict failure rapidly and safely.
     
    % based on future prediction, uncertainty estimation, and a combination of model-free acting and planning, that allows an agent to select the demonstrations that are most informative to reduce its epistemic uncertainty. We empirically demonstrate how this method increases the performance on a set of manipulation tasks by a substantial margin in both simulated and real robotics scenarios. We also show how this method enables the agent to autonomously predict failure rapidly and safely. Finally, we demonstrate how the proposed method is able to actively minimize state distribution shifting at test time, one of the main cause of erroneous and dangerous behaviors.
\end{abstract}

\section{Introduction}
To teach a robot a novel task, one promising avenue of research is \textbf{Imitation Learning} (IL) \cite{osa2018algorithmic, hussein2017imitation, ghasemipour2020divergence, zhang2018deep, bojarski2016end}, and in particular, \textbf{Behaviour Cloning} (BC). In BC, an expert collects a series of demonstrations of how to solve a task, and an agent then uses supervised learning to generalise a policy to new instances of the task. Showing how to solve a task can often be easier and faster than directly designing a controller manually, or hand-crafting a potentially complex reward function and learning a policy through reinforcement learning. However, once deployed, the agent state visitation distribution may be different from its training distribution, causing erroneous and dangerous behaviour \cite{pomerleau1989alvinn}. This may have two main causes: a poor state space visitation during the expert's demonstrations, and compounding errors from the policy at test time that cause the state visitation distribution to drift outside of the training distribution.

Therefore, we argue that a robot should both \textbf{maximize the information obtained from demonstrations}, and at test time should act in such ways as to solve the task while staying close to the states explored during training, by \textbf{actively minimizing state-distribution shift} through a feedback action. Furthermore, when out-of-distribution (OOD) states are encountered, this should be detected quickly and \textbf{execution should be autonomously stopped}.

Based on these principles, in this work we propose a novel active imitation learning and control framework that tackles the aforementioned problems: \textbf{SAFARI} (\textbf{SAF}e and \textbf{A}ctive \textbf{R}obot Imitation Learning with \textbf{I}magination). SAFARI (Figure 1) is a framework for imitation learning and control which allows the robot to (1) actively ask for more informative demonstrations at training time, (2) actively minimize the distributional shift at test time, and (3) predict test time failures to improve safety. These methods all contribute to tackle the aforementioned issues from different perspectives, resulting in a substantial improvement in performance and safety with respect to baseline methods. 
%To tackle the aforementioned problem, it is necessary to first predict state-distribution shift. Hence, we designed a system which can identify when the robot is heading towards out-of-distribution (OOD) states, by learning both a model which can predict future states and their distributional distance to the training distribution. This enables the robot to safely stop execution at test time when an OOD state is likely in the near future. In addition, this enables the robot to actively ask the expert for a demonstration at this point, which is a more data efficient framework than simply collecting demonstrations passively. Furthermore, these predictions allow the robot to actively reduce state distribution shift at test time by online planning in addition to following its policy. As we show in this work, these methods all contribute to tackle the aforementioned issues from three different perspectives, resulting in a substantial improvement in performance and safety.

Specifically, SAFARI uses an interplay between a \textbf{policy network}, a \textbf{learned dynamics network} (also referred as \textbf{world model} in this work), and an \textbf{epistemic uncertainty network}. The policy network is trained to emulate the expert's behavior with BC \cite{levine2016end, finn2016deep, zhang2018deep}, while the dynamics network \cite{nagabandi2018neural} is trained to predict future states, enabling planning alongside model-free policy acting, and the uncertainty network \cite{di2018improving, boney2019regularizing} is trained to estimate epistemic uncertainty and out-of-distribution states, similarly to an energy model \cite{lecun2006tutorial}.  %By computing a degree of familiarity with the proposed instance of the task, the agent can prevent the expert from solving instances that are familiar and hence less informative, thus being more time and sample efficient, but also to focus on instances that are particularly unfamiliar, and are hence more probable to result in a failure.
 %The interplay of these models allow the robot to ask for the most informative demonstrations at training phase, thus expanding as possible the distribution of visited states, while at test phase it emulates the expert's demon

%Our method is based on Denoising Autoencoders for epistemic uncertainty estimation \cite{vincent2010stacked, arponen2017exact, boney2019regularizing}. With "uncertainty" we will always refer to epistemic uncertainty: dealing with aleatoric uncertainty is outside the scope of this work. We show how this kind of network can successfully learn to compute an approximation of the uncertainty of the agent, and thus predict online the probability of success of the agent, in an unsupervised way. 

\begin{figure*}
    \centering
  \includegraphics[width=0.32\linewidth]{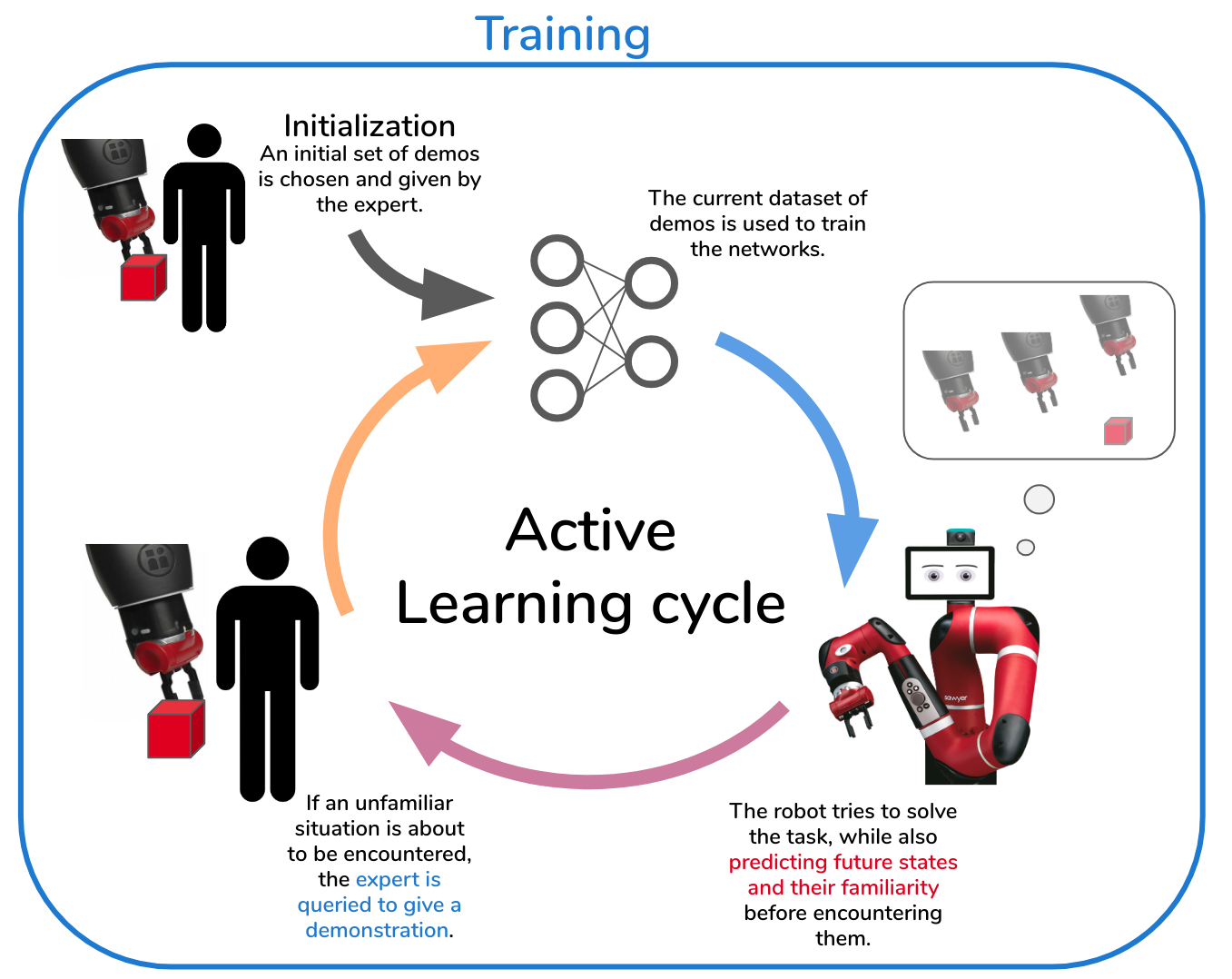}
  \includegraphics[width=0.50\linewidth]{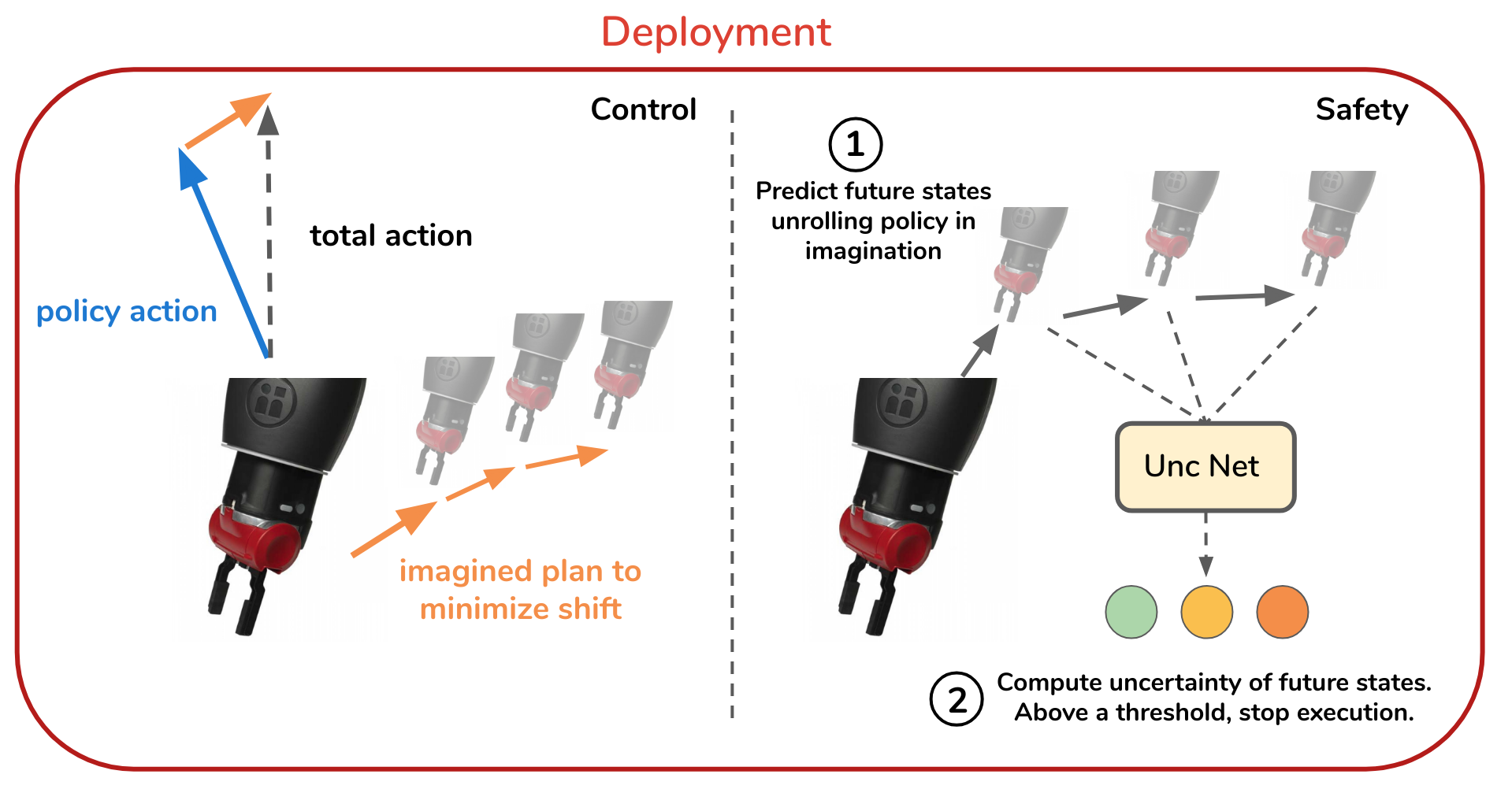}

  \caption{An overview of our training and control framework, SAFARI. We start with an active imitation learning phase, which trains the policy, uncertainty and dynamics networks. These networks are then used in deployment to minimize state-distribution shift and predict dangerous states, improving performance and safety.}
  \label{fig:boat1}
\end{figure*}

We tested our methods on both simulated and real robotics environments. For the simulated experiments, we used three robot manipulation tasks: bringing a cube to a desired position, stacking two cubes at a desired position, both with the Fetch robot, and a nut-and-peg insertion task with a Sawyer robot, all using the MuJoCo physics simulator \cite{todorov2012mujoco}. As real-world experiments, we used a Sawyer robot, with a task to move an object on a table to a goal position, using the end-effector position and an RGB stream from a camera as inputs.

To summarise, the main contributions of this work are the following:
\begin{itemize}
\item We demonstrate how our Active Learning method is able to \textbf{select the most uncertain, and hence informative, instances of the task}, in order to minimize the number of required demonstrations.
\item We propose a new hybrid policy, which \textbf{combines the model-free output} of a policy network, trained to emulate the expert's behavior, \textbf{and an online plan} that aims at \textbf{bringing the state closer to the distribution of states visited by the demonstrations}. This allows to minimize state-distribution drifting caused by compounding errors at test time.
\item We demonstrate the ability of our uncertainty estimation algorithm to \textbf{rapidly and safely predict failures at test time}, avoiding possibly dangerous situations that could harm the robot and its environment.
\end{itemize}

We bring these approaches together into a \textbf{unified framework called SAFARI}, that uses a common approach based on an interplay of a policy, a learned world model, and an epistemic uncertainty network to tackle real-world Imitation Learning from three different perspectives.
%\item We bring these approaches togetherWe propose SAFARI, a novel method for \textbf{active robot imitation learning and control}, using an interplay between a policy network, a learned world model, and an epistemic uncertainty network.

%\begin{figure}[t]
%    \centering
%  \includegraphics[width=0.24\linewidth]{robot1.png}
%  \includegraphics[width=0.25\linewidth]{robot2.png}
%  \includegraphics[width=0.24\linewidth]{robot3.png}
%  \includegraphics[width=0.25\linewidth]{robot4.png}
%  \caption{These pictures, from left to right, show an example of our iterative active learning method. The robot tries to solve an instance of the task. When its uncertainty surpasses a threshold, the human operator is called in order to demonstrate how to complete the task from that configuration. The robot collects the new demo and updates its networks.}
%  \label{fig:boat1}
%\end{figure}

\section{Related Work}
Imitation Learning with deep neural networks has shown impressive results in recent years in robotics. (\cite{argall2009survey, zhang2018deep, james2018task, gupta2019relay, james2017transferring}). %\cite{lynch2019learning, sermanet2018time, zhang2018deep, kumar2015mujoco} showed how virtual-reality can be a straightforward and quick way to provide demonstrations to the agent. \cite{levine2016end} showed the ability of deep convolutional networks to extract meaningful features from raw inputs. \cite{finn2016deep} demonstrated how to use the low-level features extracted by such network as inputs to policies and controllers. \cite{james2017transferring} showed how low-dimensional states available in a simulator can be used to generate demonstrations.

In order to obtain more informative demonstrations that properly explore the state space, one effective approach is using Active Learning algorithms \cite{argall2009survey}. Active Learning \cite{hanczor2018improving} deals with the design of methods able to actively ask an expert for labels on unseen data, improving data-efficiency with respect to passive learners.   \cite{ross2011reduction} proposed DAgger, a popular active imitation learning algorithm, to tackle state visitation shift. Several variants have been proposed over the years \cite{menda2017dropoutdagger, cronrath2018bagger, hanczor2018improving}, but these methods require the expert to label states with single, optimal actions, a method that cannot be applied efficiently to real-world systems. \cite{kelly2019hg} extends these methods by allowing a human expert to take control at training time, without querying them on single states. Our method extends these approaches by more safely and efficiently locating OOD state areas using future predictions, querying the expert at train time or stopping execution at test time without human supervision. We also compare our method with \cite{laskey2017dart}, which injects noise during the expert's demonstrations to simulate the errors of a policy in order to learn how to compensate for it, artificially expanding the visited states during demonstrations. 

To minimize state distribution shift at test time, Adversarial Imitation Learning methods such as \cite{GAIL, AIRL} have been shown to be superior to Behavioural Cloning \cite{ghasemipour2020divergence}. These methods, similarly to our approach, learn a function that can distinguish between expert's and non expert's state visitation distributions. However, a substantial difference is that these methods use model-free reinforcement learning to learn a policy that can minimize distribution shifts. This is both dangerous and impractical in real robotics scenarios, needing autonomous exploration, environment resets, and many time-steps \cite{ghasemipour2020divergence}. In our work, we designed a method that tackles the same problem, but designed to be efficiently trained and safely deployed on real robots, using only supervised data from the expert's demonstration: instead of a discriminator that learns from expert's and agent's data, we use a Denoising Autoencoder that can learn to discriminate by only receiving expert data. Then, instead of learning a policy with model-free reinforcement learning, we learn a dynamics model that allows the agent to plan online for a series of actions that can minimize state distribution shift. A more detailed mathematical comparison between these approaches can be found in section III.F.

Computing epistemic uncertainty is a crucial aspect of our work. We use a Denoising Autoencoder to learn to approximate the data distribution.
\cite{di2018improving, boney2019regularizing, zhu2017deep, arponen2017exact} demonstrated how (Denoising) Autoencoders \cite{vincent2010stacked} can be used for detecting out-of-distribution inputs and regularizing their effects on neural networks, with \cite{boney2019regularizing} comparing them to ensemble methods. \cite{boney2019regularizing} also showcases the utility of using Autoencoders to regularize plans under a learned dynamics model, by minimizing the distribution shift with the experience collected during training, which in our case is the expert's demonstrations.

\cite{filos2020can} is the closest work to our method, but applied to a very different scenario: simulated autonomous vehicles. While proposing a similar approach, they don't use a policy network and don't learn a dynamics function, but instead compute a series of desired future states. They assume the existence of an inverse dynamical model which can then deduce actions, which is unattainable in robotic manipulation with objects and visual inputs, and hence we designed a dynamics model to learn that from data.

%
%\begin{figure}[]
%  \centering
%  \includegraphics[width=0.34\linewidth]{fetchpaper.png}
%   \includegraphics[width=0.35\linewidth]{sawyerpaper.png}
%  \caption{Pictures of two of the three tasks. Left: Fetch Robot on 2 Cubes Stacking. Right: Sawyer Robot on the Nut-and-Peg insertion task.}
%  \label{fig:boat1}
%\end{figure}

%\begin{wrapfigure}{r}{0.5\textwidth}
%  \begin{center}
%    \includegraphics[width=0.50\textwidth]{cycle_full.png}
%  \end{center}
%  \caption{An overview of our active learning method.}
%\end{wrapfigure}

\section{Method}

In Imitation Learning \cite{zhang2018deep, james2018task, lynch2019learning}, a human expert gathers a set of demonstrations of how to successfully solve a task in the form of trajectories of states and actions $(s_0, a_0, ..., s_T, a_T)$. A policy network $f_\theta$ is then trained on this data, with the hope that it will generalize to novel instances of the task. As described in the Introduction, once deployed the learned policy can encounter states that are different from the dataset gathered by the expert. In these situations, the output of the parametrised policy can be erroneous and even dangerous for the robot and its surroundings \cite{pomerleau1989alvinn, ghasemipour2020divergence}. To tackle these problems, we aim at collecting the most informative demonstrations at training time through an Active Learning phase. In addition, at test time, once the policy has been trained, we aim at minimizing state visitation shift through a feedback action that leads the agent closer to the distribution explored by the expert, where the policy's behaviour is more likely to be correct. \cite{pomerleau1989alvinn}

We tackle these problems using a common approach, based on an interplay between a policy network, a dynamics network, and an uncertainty network. In the following section, we describe more in depth the aforementioned networks (III.A), then describe our approach for computing epistemic uncertainty, which is then used in three different ways, in each of the three components of our framework (III.B). Then, we introduce the Active Learning method we implemented and use at training time (III.C) and the state distribution shift minimization method we use at test time (III.D). We then describe our method for failure prediction at test time (III.E). Finally, we analyse more in depth the mathematical aspects of our methods, describing the similarities and differences with current state-of-the-art approaches \cite{ghasemipour2020divergence} (III.F).

\begin{figure}[t]
  \begin{center}
    \includegraphics[width=0.9\linewidth]{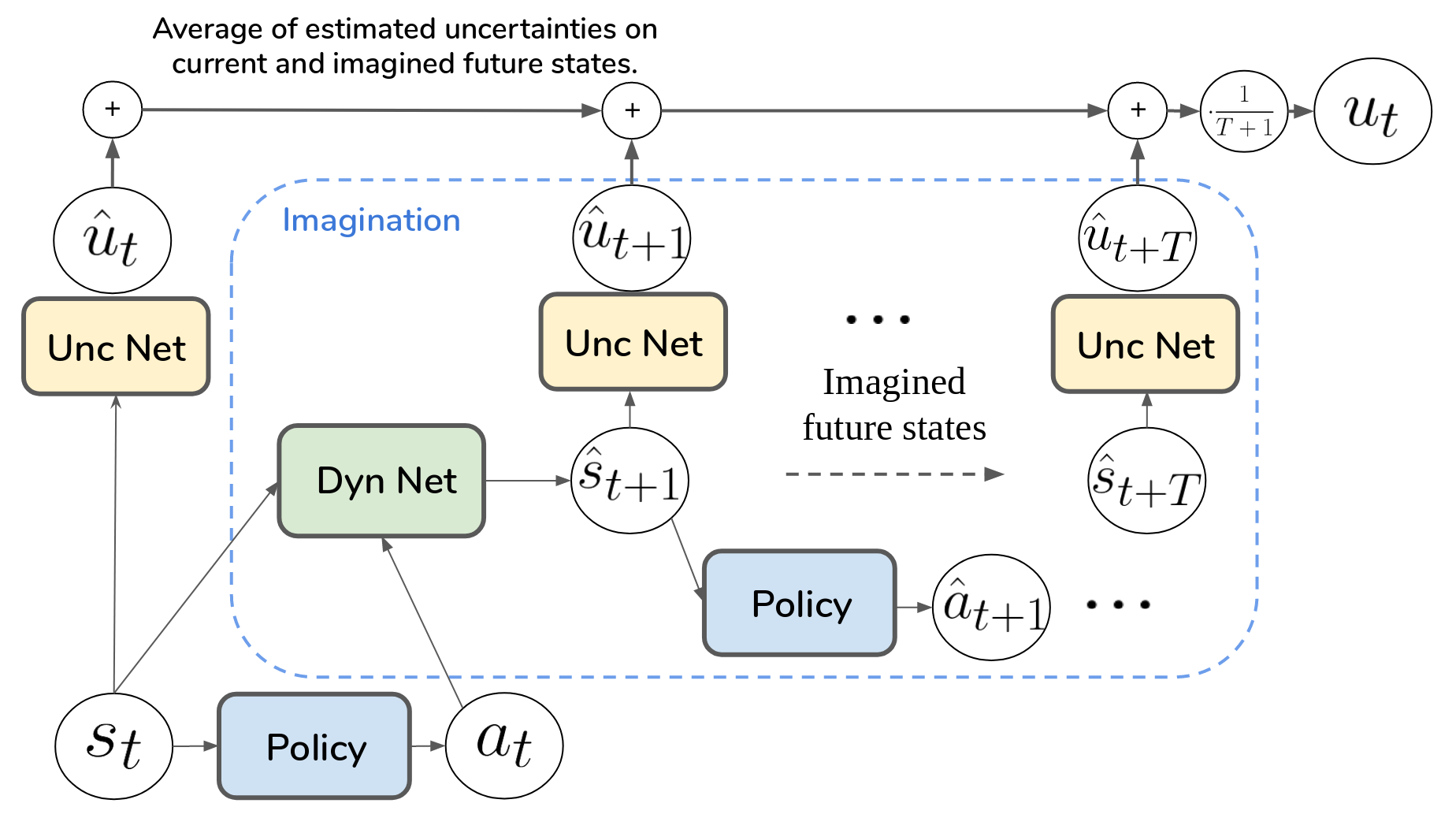}
  \end{center}
  \caption{This diagram depicts the computation graph to estimate the agent's uncertainty described in Section III.C. At each time step, the agent predicts the future visited states by running its policy network using its internal world model. It then computes the epistemic uncertainty of these imagined state using its uncertainty model, averaging the results over all $T$ steps.}
\end{figure}

\subsection{Networks Models and Architectures}
Inspired by previous works on Imitation Learning \cite{lynch2019learning, levine2016end, finn2016deep, zhang2018deep}, we use a feed-forward neural network $f_\theta$ to parametrise our policy, taking as input the current observation, and computing as output the action. The policy network is trained with Behavioural Cloning. 

To model the uncertainty of the agent, we use Denoising Autoencoders, $g_\phi$. The Denoising Autoencoder \cite{vincent2010stacked} is a feed-forward network that takes as input the current state, that during training time is corrupted with Gaussian noise, $\tilde{x}$, and is trained to give as output the denoised input, $e = \sum_i(x_i - g_\theta(\tilde{x})_i)^2$, where $e$ is the denoising error that we wish to minimize with gradient descent. The denoising error is akin to the output of an energy model \cite{lecun2006tutorial}, hence representing a proxy of the distance of the input from the training distribution: if those were part of the previous training samples, the errors will be lower, while being higher on novel states or goals \cite{di2018improving, boney2019regularizing}. In this work we will also refer to the Denoising Autoencoder as the uncertainty network. 

As a learned dynamics model, we use a feed-forward network $d_\gamma$ that takes as input the current state and action, and predicts the one-step difference $\Delta{} \hat{x}_{t+1}$ such that $\hat{x}_{t+1} = x_t + \Delta{} \hat{x}_{t+1}$ \cite{nagabandi2018neural}. 

All the networks are trained together on the same data (although used in different ways).

\subsection{Uncertainty Estimation with Autoencoders and Imagination}
The backbone of these approaches is the accurate computation of the epistemic uncertainty of the policy. In this work, we propose a novel approach at computing the familiarity of the agent with its current and also future states. When we refer to "uncertainty" in this work we will also refer to epistemic uncertainty. Dealing with aleatoric uncertainty is outside the scope of this work.

To compute the uncertainty on a particular state, we use an interplay of a policy network, a dynamics network, and an uncertainty network, as we show in Figure 2. 

From a state $s_t$, we use the policy network and the dynamics network to predict the future states that the agent will encounter following its current policy, $s_{t+1}, ... , s_{t+T}$. We use the Autoencoder $g_\phi$ to compute the aforementioned error on each of those states, and then average the result. For brevity, we will refer to the uncertainty of a state as $u_t = u(s_t, f_\theta, d_\gamma, g_\phi)$, where $s_t$ is the current state and $f_\theta, d_\gamma, g_\phi$ are the policy, dynamics and uncertainty networks, that are used as showed in Figure 2. This value is used as a proxy for the epistemic uncertainty of the policy on the current state.
%This estimation is used in all our three proposed methods: 1) at training time, it allows the robot to \textbf{query the expert to complete the task} if the uncertainty estimation is over a threshold (III.C). At test time, the epistemic uncertainty on future states is used to \textbf{compute a series of actions that will minimize this value}, hence also minimizing distributional shift (III.D). Finally, if this value becomes larger than a threshold at test time, \textbf{a failure is predicted and execution is stopped}, strongly improving safety for the robot and its surroundings.
In section IV, we empirically show how this technique can be used to gather the most informative demonstrations at training time (IV.A), and to minimize state distribution shift (IV.B) and eventually predicting a failure (IV.C) at test time.
%We show empirically in Section IV.C how this method can accurately predict failures several steps before they happen, surpassing even a supervised model, allowing to collect informative demonstrations at training time and adding safety at test time.

\begin{figure}[t]
    \centering
  \includegraphics[width=0.44\linewidth]{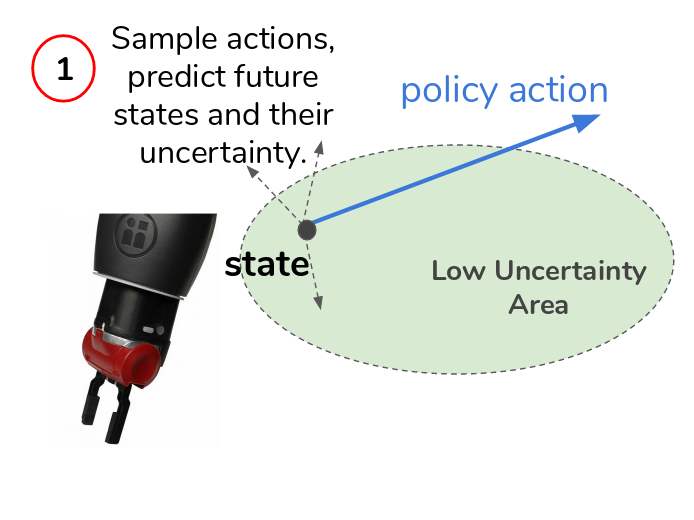}
  \includegraphics[width=0.53\linewidth]{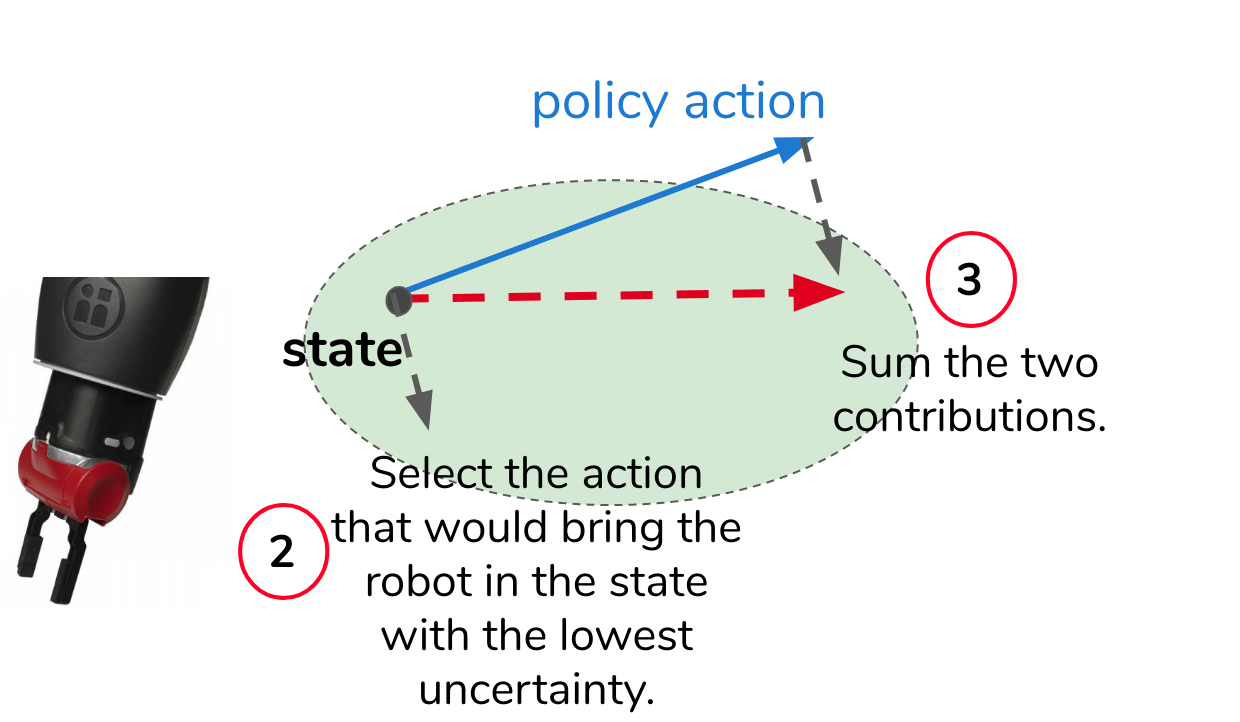}
  
  \caption{In this figure we visually describe the hybrid behavior of the robot that we propose. After computing the output of the policy network, the agent plans for a series of actions that will minimize state distribution shift as described in Section III.D, and then combines the two contributions.}
  \label{fig:boat1}
\end{figure}

\subsection{Active Imitation Learning with Uncertainty Estimation}
In many Imitation Learning works \cite{zhang2018deep, james2018task, lynch2019learning}, the demonstrations collection phase is separated from the robot interaction with its environment. We propose instead to have an \textbf{iterative, interactive approach to gather demonstrations}.
Instead of receiving all the demonstrations beforehand, the agent receives only a small set of demonstrations as an initialization phase. The agent can then request a demonstration by interacting with its environment, trying to solve the task following its policy network and stopping when its uncertainty surpasses a certain threshold. Each new demonstration is then added to the training data and the networks are retrained. The uncertainty estimation method is described in section III.B. Our experiments, detailed in Section IV, demonstrate
how we can train a better policy network using active demonstrations than receiving the same number of demonstrations passively beforehand. %Instead of receiving all the demonstration beforehand, the robot receives a small set of initial demonstration. It then learns to emulate the expert's behavior, while also learning an uncertainty model trained on the states distribution visited by the expert. The agent will now interact with its environment and ask for demonstrations when its uncertainty surpasses a certain threshold. The demonstration is then added to the dataset of trajectories $D$. Our experiments, detailed in Section IV, demonstrate how, with the same number of demonstrations, our method obtains substantially better performance at test time.

%\begin{figure}[t]
%    \centering
%  \includegraphics[width=0.40\linewidth]{fetchpaper.png}
%   \includegraphics[width=0.31\linewidth]{sawyer.png}
%  \caption{We tested our method on three tasks, two in a simulated robotics environment (left) and one on %a Sawyer robot (right). The tasks are, respectively, to pick and place a cube to a goal position, stacking %two cubes one on top of the other, and moving an object on a table to a goal position.}
%  \label{fig:boat1}
%\end{figure}

\subsection{Online Planning for State Shift Minimization}
Here we demonstrate how an interplay of the networks we introduced in this work can be used to tackle state distribution shift at test time, one of the most important causes of failures of Imitation Learning \cite{ross2011reduction, pomerleau1989alvinn, laskey2017dart, ghasemipour2020divergence}. While our method allows to quickly anticipate these scenarios and hence predict failure (as we empirically show in Section IV.C), we want to prevent this from happening altogether when possible and maximize the number of successfully completed tasks.
At test time we combine model-free actions given by the policy to solve the task with actions planned by predicting the future to minimize the robot's epistemic uncertainty (Figure 3). More specifically, at test time the agent plans for a series of actions $(a_0, a_1, ..., a_T)$ that minimize the uncertainty of the predicted future state visited by applying these actions, $ \hat{s}_{t+T+1}$. The uncertainty of this state is computed as described in Section III.B. Hence, at each time step the robot computes its action as
\begin{multline}
 a_t = f_{\phi}(s_t) + \beta \cdot a_{p,0}  \text{ where } a_{p,0:T} = argmin_a
 \text{ } u(\hat{s}_{t+T+1} ) \\ \text{ subject to } \hat{s}_{t+1} = d_{\gamma}(s_t, a_t) 
%\[  \]
\end{multline}
where $a_{p}$ is a trajectory of actions that minimizes the uncertainty of the final predicted state, $T$ is the length of the plan, and $a_{p,0}$ is the first action of such trajectory, following the Model Predictive Control framework.
Intuitively, we are combining a policy function with an online planning phase that outputs the action that is predicted to bring the robot to a more familiar state, defined as being closer to the state's visitation distribution encountered in the expert's demonstrations, hence increasing the probability of a successful task completion. The hyperparameter $\beta$ is used to tune the influence of the planned actions. High values of $\beta$ result in more conservative behaviour. We found $\beta = 0.2$ to give the best performance in our tests. Our experiments, described in Section IV, demonstrate how this approach improves performance on robot manipulation tasks with respect to only using the policy.

\subsection{Failure Prediction at Test Time}
At test time the agent computes, at each time step, its epistemic uncertainty on the current and future predicted states (Figure 2) as described in III.B. This allows the robot to predict if it is headed towards an unfamiliar area of the state space, where the robot is likely to fail the task, and even damage itself or its environment. While we actively try to minimize this value at test time (III.D), if the estimated uncertainty becomes larger than a threshold, the agent stops execution. In our experiments section, we show how this combination of an uncertainty network and future prediction through policy and dynamics networks allows not only to accurately predict failures, but also to do so rapidly.

\subsection{A Mathematical Analysis of Imitation Learning as Divergence Minimization}
As described in \cite{ghasemipour2020divergence}, Imitation Learning can be framed as a divergence minimization problem between the expert's state-action visitation distribution, $p_{exp}(s,a)$, and the agent's. In Behavioral Cloning, as in our work, a policy $\pi(a | s)$ is trained with supervised learning to minimize $\text{KL}_{p_{exp}(s)}(\pi_{exp}(a|s)||\pi(a|s))$, that becomes a regression problem to minimize $-\E_{p_{exp}(s)}[\log(\pi(a|s))] $. While being a sample-efficient model, by only imitating the expert's action, a policy may fail when facing OOD states \cite{ghasemipour2020divergence}. Other methods, like GAIL \cite{GAIL}, use model-free reinforcement learning to emulate not only the action, but the full state-action visitation distribution of the expert, by minimizing $ \text{D}_{JS}  (p_{\pi}(s,a)||p_{exp}(s,a))$. In this case, the agent autonomously explores the environment: a discriminator, $D(s,a)$, is trained on the expert's and agent's trajectories to discriminate between expert and policy state-action pairs \cite{GAIL}. The agent is then rewarded if the discriminator classifies the state-actions it visits as belonging to the expert's distribution. While obtaining strong performance \cite{ghasemipour2020divergence}, this and similar methods \cite{AIRL} are based on model-free RL and require possibly unsafe autonomous exploration from the agent. In our work, we train an uncertainty model on the expert's state distribution, which can learn in an unsupervised way to discriminate between the expert's and agent's state-visitation distribution (IV.B). We then use online planning with this learned model to minimize the divergence between the agent's current state and expert's state-visitation distribution. Hence, we obtain the same effect as \cite{GAIL, AIRL, ghasemipour2020divergence} but leveraging only supervised data, hence being substantially safer, in the form of the expert's actions and trajectories of states used to learn a dynamics model.

%\footnote{Code and videos can be found at the anonymized site \texttt{https://sites.google.com/view/robotactivelearning/}.}

\begin{figure}[t]
  \begin{center}
    \includegraphics[width=0.95\linewidth]{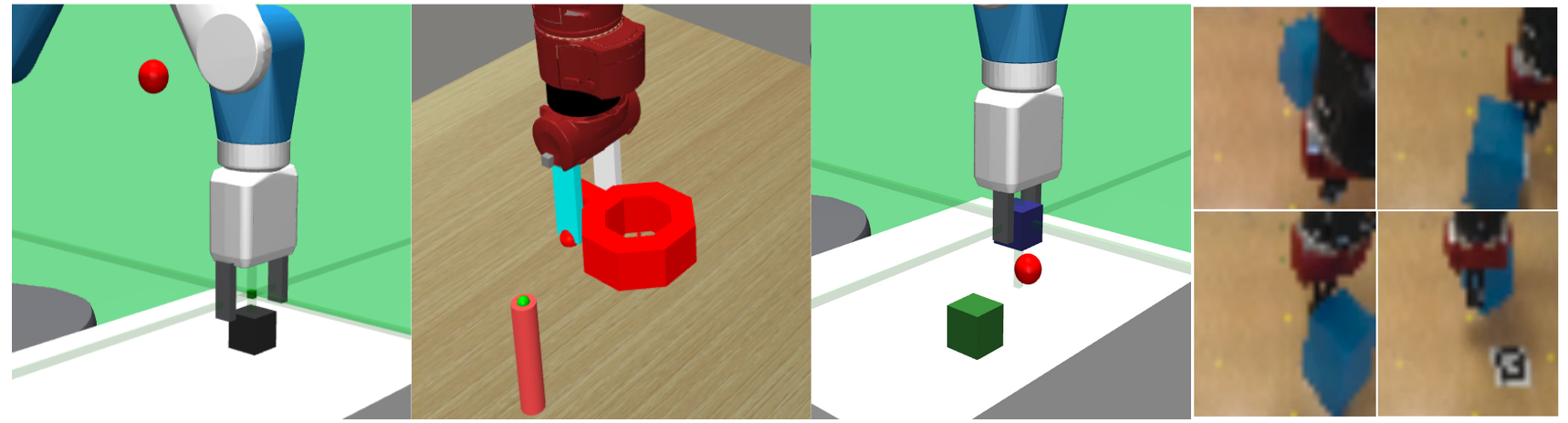}
  \end{center}
  \caption{The simulated and real environments used in this work.}
\end{figure}

\section{Experiments}
In this section, we describe the experiments we designed to benchmark the performance of SAFARI on several manipulation tasks, both simulated and on a real robot (Figure 4). These experiments have been designed to answer the following questions \footnote{Detailed hyperparameters, code and videos can be found at the following link: \texttt{https://www.robot-learning.uk/safari}}:

\begin{itemize}
\item Does our proposed active imitation learning method perform better at test time than the baselines \textbf{given the same number of demonstrations}? (IV.A)
\item Does online planning to reduce state-distribution shift \textbf{improve performance at test time with respect to only using the policy network}? (IV.B)
\item Can our method \textbf{reliably predict failure} during execution before it happens? Does imagination \textbf{reduce the number of steps needed to predict failure}? (IV.C)

\end{itemize}

%Further discussion, ablation studies and a more detailed list of hyperparameters used in all the experiments can be found in the Appendix.

\subsection{Active Learning Performance at Test Time}
We designed the following experiments to measure the ability of our method to gather informative demonstrations, efficiently exploring the state space at training time. We compare our proposed Active Learning method against several baselines: Passive Learning (PL), randomly stopping during execution to ask for a demonstration (Rand On Policy), and DART \cite{laskey2017dart}, an effective algorithm for robot imitation learning that tackles distribution shift.

$N$ demonstrations are used to train a policy network, consisting of our Passive Learning baseline. For Active Learning, we select the initial $(1 - \gamma)N$ (where $0 < \gamma < 1$) demonstrations, used as the initialization demonstrations, and then collect an additional $\gamma N$ demonstrations using our iterative method. We enforce that both $(1-\gamma)N$ and $\gamma N$ are integers. 

%For Active Learning, we follow these steps: we train the Policy Network, the Dynamics Network and the Uncertainty Network on the $N$ demonstrations. The expert can then move the initial configuration of state and desired goal until the computed uncertainty is above a threshold, or compute the uncertainty for $k$ initial configurations and then select the highest one. The expert then provides a demonstration for the selected initial configuration. It repeats these steps $m$ times, after which the networks are retrained. These steps are repeated until the expert has gathered additional $N$ demonstrations. Hence, in Passive Learning and Active Learning we use the same number of demonstrations, $2N$.

%During the Active Learning part we follow these steps: we train the Policy Network, the Dynamics Network and the Uncertainty Network on the current set of demonstrations. Then, we initialize an instance of the task and let the agent try to solve it, whilst measuring its uncertainty online at each step. If the uncertainty becomes greater than a threshold, the agent autonomously stops and asks for a demonstration on how to solve the task starting from where it stopped.
We collected the Active Learning demonstrations following the method described in III.C. The uncertainty threshold is computed as $u_{thr} \cdot \text{err}_{train}$, where $u_{thr} > 1$ is a hyperparameter, and $\text{err}_{train}$ is the average uncertainty on the training set computed after training the uncertainty network $g_\phi$. We found the method to be robust to changes to this hyperparameter. Each demonstration trajectory was then added to the training data and the networks were retrained. We iterated until we collected $\gamma N$ additional demonstrations, for a total of $N$ demonstrations. We then tested the performance of the two policy networks, trained using the passive and active demonstrations, on the $M$ test configurations of the task. 

For the Random On-Policy benchmark, we collected $N$ demonstrations by letting the robot follow its policy, then stopping execution at a random step and asking the expert to complete. This benchmark is designed to measure if our method is actually able to recognize useful demonstrations to obtain, or if the only advantage comes from on-policy stopping.

For DART, we collected $N$ demonstrations following \cite{laskey2017dart}, by injecting noise during the expert's demonstrations. We followed the original authors' code to implement this baseline.

An \textbf{experiment} is defined as a complete round of data collection, training, and testing, all with one random seed, which is different for each experiment. Each experiment involved testing over a different test set, composed of a number of different initial conditions, and recording the number of successes. Then, each entry in Table I shows on what percentage of experiments each of the two compared methods outperformed the other in terms of the number of successes in that experiment. The total, normalized number of successfully completed test instances can be found in Figure 5, B. The reason why we show these results both as in Table I and in the raw success rates (Figure 5, B), is that we believe this is a clearer and more statistically significant way of directly comparing two methods with respect to robustness over random seeds.

\subsubsection{Simulated Environments - Cube Manipulation and Nut-and-Peg Insertion}
We start by describing our results on three simulated tasks. The three manipulation tasks are 1) 1 Cube pick-and-place, 2) 2 Cubes stacking, and 3) a Nut-and-Peg insertion task (Figure 4). The observation vector, in the simulated experiments, is composed of the pose of the end-effector, the objects, and the goal. In the real-world setting, we show how this method also scales to having vision as input. Result of Table I and Figure 5, B, are gathered over more than 7 random seeds (details can be found on our website, including the source code). In Figure 5, A, we also show how increasing the ratio of AL demonstrations increases performance. 

%\begin{table}[t]
%\centering
%\begin{tabular}{ | m{4cm} |  m{2.cm} | m{2.cm} | m{2.cm}| m{1.7cm} |} 
%\hline
%\textbf{Task name / Method} &  \textbf{AL0-PL}  & \textbf{AL5-PL} & \textbf{AL0-Rand On Policy} & %\textbf{AL0-DART} \\
%\hline
%Fetch 1-Cube Pick and Place & \textbf{22}-3 & \textbf{21}-4  & \textbf{8}-2  & \textbf{8}-2\\ 
%\hline
%Fetch 2-Cubes Stack &  \textbf{10}-2 & \textbf{10}-2 & \textbf{8}-3 & \textbf{8}-4\\ 
%\hline
%Sawyer Nut and Peg  &  \textbf{6}-2 & \textbf{7}-1  & \textbf{5}-3 &  \textbf{5}-2 \\ 
%\hline
%
%\end{tabular}
%\caption{Experimental results of Active Learning (AL) against several baselines as described in Section 4.1.1. We show how many times, over several independent runs, each method obtained a better performance on the test set. AL outperforms all baselines in every environment.}
%\end{table}

\begin{figure*}[h]
  \centering
  \includegraphics[width=0.27\linewidth]{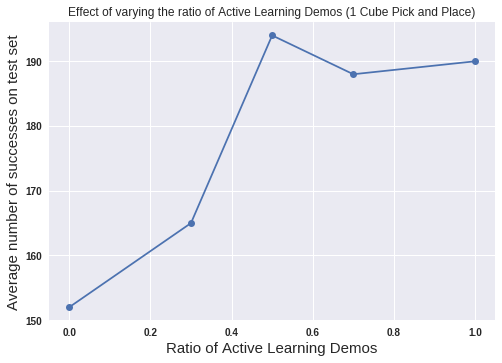}
  \includegraphics[width=0.28\linewidth]{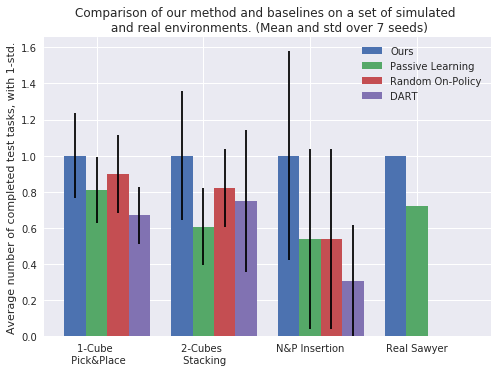}
  \includegraphics[width=0.29\linewidth]{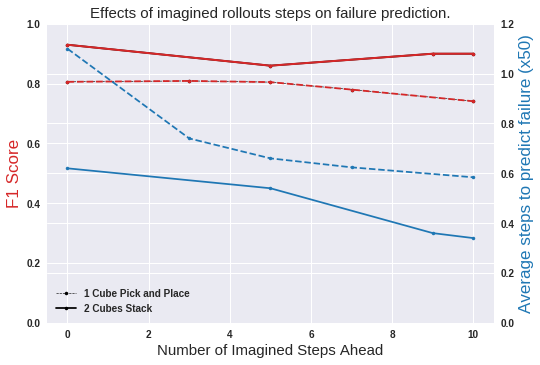}

  \caption{A (Left): How test-time performance varies as we modify the ratio of Active Learning examples over Passive Learning ones, $\gamma$. The plot represents the average number of successes over 13 different independent runs. B (Center): We show F1 score at predicting failures (red) and steps needed to predict failures (blue) on two tasks (full and dashed lines). We increase the number of imagined states from left to right. C (Right): Normalized number of test instances solved by each method, averaged over 7 seeds.}
  \label{fig:boat1}
\end{figure*}

\begin{table}[t]
\caption{}
\centering
\begin{tabular}{ | m{3cm} |  m{1.2cm} | m{1.2cm} | m{1.2cm}| m{1.7cm} |} 
\hline
\textbf{Task name / Method} &  \textbf{AL-PL}  & \textbf{AL-Rand On Policy} & \textbf{AL-DART} \\
\hline
Fetch 1-Cube Pick and Place & \textbf{88\%}-12\% &  \textbf{80\%}-20\%  & \textbf{80\%}-20\%\\ 
\hline
Fetch 2-Cubes Stack &  \textbf{83\%}-17\% &  \textbf{73\%}-27\% & \textbf{67\%}-33\%\\ 
\hline
Sawyer Nut and Peg  &  \textbf{75\%}-25\% &  \textbf{63\%}-37\% &  \textbf{71\%}-29\% \\ 
\hline

\end{tabular}
\begin{tablenotes}
\small
\item Experimental results of our Active Learning approach (AL) against several baselines. We show the percentage of experiments, defined as a round of data collection, training and testing with a different random seed (Section IV.A), on which one method successfully completes more test instances of the task than the other.
\end{tablenotes}
\end{table}

%\begin{wraptable}{r}{0.5\textwidth}
\begin{table}[h]
\caption{}
\centering
\begin{tabular}{ | m{2.3cm} |  m{1cm} | m{1cm} | m{1cm} | } 
\hline
\textbf{Active / Total Demos} &  \textbf{0/50 (PL)}  & \textbf{25/50}  & \textbf{40/50} \\
\hline
Sawyer - Object Manipulation w/ Vision & 15/25 & 18/25 & 20/25\\ 
\hline
\end{tabular}
\begin{tablenotes}
\small
\item Experimental results of different ratios of Active Learning demonstrations in a real robotics environment, described in Section IV.A.2 (PL being Passive Learning). We show the number of successfully completed tasks over the same test set of 25 tasks.
\end{tablenotes}

\end{table}
%\end{wraptable}

\subsubsection{Real Sawyer Environment}
To test the ability of our method to scale to real-world scenarios, we designed a manipulation task using the Sawyer robot. In this experiment, the goal was to slide a blue object on a table to a goal position, while having vision as an input to localize both the object and the goal, as well as the end-effector position and velocity given by internal sensors readings. The images, provided by a camera, are resized to 32x32. We use a Convolutional Autoencoder to reduce the dimensionality of the input. We gathered around 10 minutes of random movements and interaction with the environment to quickly collect an images dataset. We then trained the Autoencoder on this dataset and use it, with frozen weights, in all our experiments. The encoder transforms the input images to vectors of size 16. This vector is then concatenated to the internal sensor readings of the robot, providing the pose of the end-effector.

In this experiment, we provided a total of 50 demonstrations to the robot. Demonstration are given through kinesthetic teaching. We sampled the pose of the end-effector at 10Hz and use the one-step difference in position to compute the velocity at each time step. We then trained the policy to output these velocities given, as input, the encoded state as described before.
For the active learning approach, we tested two different ratios of active demonstrations, $\gamma = 0.5$ and $\gamma = 0.8$. 
As shown in Table 2, the active learning approach surpasses the passive learning approach in this environment as well. We also notice how a larger ratio of active demonstrations achieves the best results, as also happened in the experiments in Figure 5, A. This validates our hypothesis that demonstrations collected using our method are generally more informative and useful for the policy network to generalize.

\subsection{Test Performance of Hybrid Model-Free Policy and Online Planning}
In this section, we describe the results of our experiments designed to measure the influence of the planning method proposed in section III.D (Figure 3) on the performance of the robot at test time. We use the same environments and tasks described in section IV.A.1. 

%\begin{wraptable}{r}{0.5\textwidth}
\begin{table}[t]
\caption{}
\centering
\begin{tabular}{ | m{2cm} |  m{1.8cm} | m{1.8cm} |} 
\hline
\textbf{Task / Method} &  \textbf{SAFARI (ours)}  & \textbf{Behavior Cloning}\\
\hline
1-Cube Pick and Place & \textbf{8.8} $\pm$ 2.35 & 7.6 $\pm$ 2.35 \\ 
\hline
2-Cubes Stack &  \textbf{5.9} $\pm$ 4.0 & 3.4 $\pm$ 2.6 \\ 
\hline
\end{tabular}
\begin{tablenotes}
\small
\item Comparison of our proposed control method (Section III.D) and Behavioural Cloning (BC). Average number and standard deviation of solved test instances out of 50 test cases, averaged on 10 different seeds. In both cases, adding imagination-based state shift minimization improves performance. 
\end{tablenotes}
\end{table}
%\end{wraptable}

We collected a series of expert demonstrations of how to solve the tasks and sampled a test set. With this data, we trained the models described in this work, $f_\theta, g_\phi, d_\gamma$. We measured the performance of model-free acting \cite{lynch2019learning, zhang2018deep} on the test set by letting the robot compute actions at each time step using the policy network, $a_t = f_\theta(s_t)$. Then, we tested the performance of our hybrid acting approach on the same test set by adding, at each time step, the action computed by the policy to the first action of the uncertainty minimizing plan described in detail in section III.D.   In order to average the effect of any source of randomness (networks initialization, objects position, etc.) we repeat our experiments on 10 different random seeds, each having a different test set of 50 task instances. As shown in Table III, our proposed hybrid approach \textbf{outperforms the model-free method}.

%\begin{figure}[]
%    \centering
%  \includegraphics[width=0.20\linewidth]{inputrobot1.png}
%  \includegraphics[width=0.20\linewidth]{inputrobot2.png}
%  \includegraphics[width=0.20\linewidth]{inputrobot3.png}
 % \caption{Examples of the camera input received by the agent. This RGB stream, in addition to the position of the end-effector computed by internal sensors, compose the observation received by the robot at each timestep.}
%  \label{fig:boat1}
%\end{figure}

%\begin{figure}[]
%    \centering
%  \includegraphics[width=0.30\linewidth]{embedding.png}
%
%  \caption{Embedding of the robot's visual input and concatenation with internal states, as described in Section 4.1.2.}
%  \label{fig:boat1}
%\end{figure}

\subsection{Unsupervised Failure Prediction Performance}
In this section, we experimentally demonstrate the ability of our method to predict failures, while also minimizing the steps needed to do so. %We answer these questions: \textbf{is our epistemic uncertainty model able to predict failures? Does the dynamics model help predicting failures in fewer steps, or does it degrade performance?}

We trained our networks on a set of expert demonstrations, and we used the technique described in Section III.E to anticipate possible failures. We measured the F1 score at predicting failures of our method, along with the number of steps needed, when changing the number of future states imagined. We empirically \textbf{demonstrate how the future prediction allows our method to predict failure in fewer steps without degrading performance}. As shown in Figure 5, C, the F1 score is roughly unchanged from 0 to 10 steps of future states prediction, hence imagining future states with the dynamics model doesn't hinder failure prediction performance in that range of future steps, while the steps needed to actually predict failures are much fewer, hence \textbf{strongly improving safety}.

%remove?
%Furthermore, as a benchmark we compared our unsupervised uncertainty model, an Autoencoder, against a supervised failure predictive network, designed as a discriminator. We show that our model has comparable performance to a supervised method, trained on a dataset of failures and successes, while being trained only on the expert's demonstrations. %The details of the experiments and results are described in the \textit{A Comparison of Failure Prediction Performance} section in the Appendix.

%In our experiments, we also compared the Denoising Autoencoder with Dropout-based Bayesian approximation \cite{gal2016dropout} and Random Network Distillation \cite{burda2018exploration}, and the Denoising Autoencoders gave the best results.

\section{Conclusion}
In this work we introduced SAFARI, a framework that tackles some of the principal issues of Imitation Learning in robotic scenarios: the need for the most informative demonstrations at training time, minimization of state distribution shifting at test time, and accurate and timely prediction of failures when the robot is deployed.
We demonstrated how to tackle these problems using a common, versatile approach, obtained with a combination of policy, dynamics, and uncertainty networks, both in simulated and real robotics environments. %We plan to further explore this research avenue, that has shown very promising results regarding demonstration-efficient imitation learning, failure prediction, and improved policy performance. %\textit{Add this?: We also plan to conduct further experiments on real robotics tasks, but were unable to further do due to the COVID-19 pandemic happened in the first months of 2019.}

\end{document}